\documentclass[10pt, a4paper]{article}

\usepackage{microtype}

\usepackage{times}
\usepackage{latexsym}
\usepackage{tabularx}
\usepackage{caption}
\usepackage{subcaption}
\usepackage{graphicx}
\usepackage{paralist}
\usepackage{booktabs}

\usepackage{enumitem}

\newcounter{HWNumberOfComments}
\stepcounter{HWNumberOfComments}

\usepackage[]{lrec-coling2024} 

\title{ChatGPT Rates Natural Language Explanation Quality Like Humans: \\But on Which Scales?}

\name{Fan Huang$^1$, Haewoon Kwak$^1$, Kunwoo Park$^{2,3}$, Jisun An$^1$} 
    \address{$^1$Luddy School of Informatics,
Computing, and Engineering, Indiana University Bloomington, IN, USA\\
    $^2$School of AI Convergence, Soongsil University, Seoul, South Korea\\
    $^3$Department of Intelligent Semiconductors, Soongsil University, Seoul, South Korea\\
             huangfan@acm.org, haewoon@acm.org, kunwoo.park@ssu.ac.kr, jisun.an@acm.org\\
         }

\abstract{
As AI becomes more integral in our lives, the need for transparency and responsibility grows. While natural language explanations (NLEs) are vital for clarifying the reasoning behind AI decisions, evaluating them through human judgments is complex and resource-intensive due to subjectivity and the need for fine-grained ratings. This study explores the alignment between ChatGPT and human assessments across multiple scales (i.e., binary, ternary, and 7-Likert scale). We sample 300 data instances from three NLE datasets and collect 900 human annotations for both \emph{informativeness} and \emph{clarity} scores as the text quality measurement. 
We further conduct paired comparison experiments under different ranges of subjectivity scores, where the baseline comes from 8,346 human annotations.
Our results show that ChatGPT aligns better with humans in more coarse-grained scales.
Also, paired comparisons and dynamic prompting (i.e., providing semantically similar examples in the prompt) improve the alignment. 
This research advances our understanding of large language models' capabilities to assess the text explanation quality in different configurations for responsible AI development.
\\ \newline  }

\begin{document}

\maketitleabstract

\section{Introduction}

Artificial intelligence (AI) is progressively becoming an integral part of our daily lives, 
emphasizing the need for transparent~\cite{saxon-etal-2021-modeling, wu-etal-2023-transparency} 
and responsible~\cite{bergman-diab-2022-towards} 
AI systems. 
An essential element in achieving transparency and building trust between such systems and users is the generation of natural language explanations (NLE)~\cite{kumar-talukdar-2020-nile}. 
The NLEs play a crucial role in clarifying the reasoning behind AI decisions. 
As the significance of NLEs continues to grow, it has become increasingly important to evaluate the quality of these explanations~\cite{yao-etal-2023-human}. 

Traditionally, evaluating NLEs has largely relied on gathering human judgments~\cite{clinciu2021study, yao-etal-2023-human}.
Assessing text quality through human evaluation is a crucial yet intricate endeavor~\cite{van-der-lee-etal-2019-best, yao-etal-2023-human}.
This complexity arises from two key factors: the inherently subjective nature of human text quality assessments~\cite{yao-etal-2023-human} and fine-grained ratings on a Likert scale~\cite{van-der-lee-etal-2019-best}. 
{Furthermore, it is challenging to eliminate unintended biases in question wording~\cite{schoch-etal-2020-problem} or participant recruitment~\cite{kwak2022who} in collecting human responses.}
Consequently, human evaluation can be resource-intensive and time-consuming. 
Developing models capable of autonomously assessing explanation quality could be a valuable complement to human evaluations, which is a critical step toward 
responsible AI systems~\cite{chiang-lee-2023-large}.  

The emergence of the new generation of large language models (LLMs), such as InstructGPT~\cite{ouyang2022training} and ChatGPT~\cite{openai_chatgpt}, has demonstrated remarkable ability in understanding natural language. These models have leveraged extensive knowledge accrued during training to outperform prior approaches in various tasks, including open-domain QA, document summarization, and mathematical reasoning~\cite{qin2023chatgpt, wang2023robustness, bang2023multitask}.
ChatGPT has also exhibited human-level competency in generating informative and clear NLEs, especially in contexts like hate speech detection~\cite{IsChatGPT}.
This progress naturally leads to whether LLMs can evaluate the quality of explanations. 
As AI-driven systems play pivotal roles in applications where explaining their decisions is imperative, ensuring the accuracy and alignment of LLM's assessments with human judgments becomes increasingly essential.
While numerous studies have investigated the potential of LLMs to replace or augment human annotations, the primary focus has been on classification tasks such as topic and stance detection~\cite{Yi_Liaw_Huang_Benevenuto_Kwak_An_2023, gilardi2023chatgpt,IsChatGPT}, 
with little attention given to their ability to assign ratings or distinguish among ordinal categories, a focus of our study.

In this study, we delve into the alignment between ChatGPT's evaluation of explanation quality and human assessments using three distinct datasets: e-SNLI for logical reasoning~\cite{e-snli}, LIAR-PLUS for misinformation justification~\cite{liar_2018}, and Latent Hatred for implicit hate speech explanation~\cite{elsherief-etal-2021-latent}. 
We engage human annotators and ChatGPT to assess human-written explanations, focusing on two widely used metrics, informativeness and clarity~\cite{gatt2018survey,howcroft2020twenty,clinciu2021study}, which are in 7-point Likert scale. 
We employ various experiments, framing the problem as a classification task and pairwise comparisons as they can be easier~\cite{NIPS2006_trueskill} and reduce biases and errors~\cite{narimanzadeh2023crowdsourcing} for human annotators, to evaluate ChatGPT's capability to approximate human evaluations of explanation quality. 
Moreover, given the promising advantages of the in-context learning prompting approach~\cite{xie2021explanation}, we investigate the potential benefits of dynamic prompting, in which we provide customized instruction with the most semantically similar examples.

Our findings reveal that ChatGPT's evaluations closely approximate human assessments in coarse-grained (binary and ternary classifications) but encounter challenges in fine-grained (7-way classification) assessments. Notably, ChatGPT excels in comparative evaluations akin to human judgment. The introduction of dynamic prompting shows promise in improving results, highlighting ChatGPT's potential to reduce the cost of data annotation for subjective metrics. 
This research contributes to our understanding of LLM's capacity to evaluate the explanations, adding valuable insights for ChatGPT's potential role in replacing or supplementing manual evaluations when building responsible AI systems.\footnote{Our code and data are publicly available at ~\url{https://github.com/muyuhuatang/ChatGPTRater}.}

\section{Related Works}
\label{related-works}

\noindent \textbf{Natural Language Explanation}
Advanced language models like GPT-3~\cite{brown2020language} and its successors 
have already shown the great capability of providing natural language explanations~\cite{wang2023evaluating}.  For example, \citet{IsChatGPT} find that ChatGPT can generate human-level quality text, and people prefer ChatGPT-generated text to human-written text in terms of informativeness and clarity evaluation perspectives. Also, studies have shown that it is challenging to discern ChatGPT-generated text from human-written text reliably~\cite{sadasivan2023can}.

In the complex environment of evaluating advanced LLMs, choosing a tool that provides depth and precision of analysis is critical. Natural language explanation (NLE), with its inherent complexity and nuance, provides a vital avenue to perceive better the limitations and capabilities of advanced LLMs~\cite{e-snli, liar_2018, clinciu2021study, elsherief-etal-2021-latent}. The nuanced features of NLEs allow for finer-grained evaluation from various aspects~\cite{cambria2023survey}, providing the fertile foundation for in-depth study of the multi-dimensional capabilities of LLMs from understanding linguistic subtleties to assessing cognitive consistency in generated responses~\cite{e-snli, emelin-etal-2021-moral, sakai2021framework}. 

More human-preferred NLEs are generated through the ChatGPT, compared with the collected ground truth human annotations explaining the implicit online hate speech~\cite{IsChatGPT}. The MTurk workers also use ChatGPT to provide textual annotations, and it is hard for the task requester to distinguish the human-written and machine-generated textual annotations~\cite{veselovsky2023artificial}. A handful of studies investigated how well those advanced LLMs can replace human annotation results~\cite{gilardi2023chatgpt,zhu2023can}.

\vspace{2mm}
\noindent \textbf{Text Quality Evaluation}
Despite its importance, the evaluation of text quality remains not comprehensively studied~\cite{clinciu2021study}. 
The text quality evaluation generally falls into two categories: automatic metrics and human evaluation metrics. 
The automatic quality evaluation approach is based on calculating word overlap or semantic similarity between generation results and ground truth references, like BLEU~\cite{papineni-etal-2002-bleu} and BERTScore~\cite{zhang2019bertscore} metrics. 
However, they are not always correlated with human evaluations~\cite{clinciu2021study}. 
On the other hand, various human evaluation metrics have been proposed for text quality. \citet{emelin-etal-2021-moral} uses Coherence and Plausibility to measure the quality of explanations. \citet{forbes-etal-2020-social} applies Relevance to measure whether the statements about social norms correlate well with the provided contextual information considering the social norms. 
The Valid, Satisfactory, and Shortcoming scores measure the natural language description quality under the few-shot out-of-domain scenarios~\cite{yordanov2021few}. However, those metrics are domain-specific~\cite{sakai2021framework,forbes-etal-2020-social} or application-specific~\cite{emelin-etal-2021-moral, chiyah-garcia-etal-2018-explainable}.

\section{Data and Annotation}

To properly evaluate how well ChatGPT can judge the quality of explanations, it is important to test it with complex NLEs that might be challenging for the general public to understand. By examining how ChatGPT assesses the quality of explanations in these intricate scenarios, we can determine if the model can approximate human-like understanding and judgment.

We sample data instances from three existing NLE benchmark datasets. We then evaluate the quality of the explanations in the selected data instances using two widely used metrics:
informativeness~\cite{novikova-etal-2018-rankme} and clarity~\cite{van-der-lee-etal-2017-pass}. Informativeness measures how relevant the information in the explanation is for understanding the underlying reasoning or justification, and clarity assesses how clearly the ideas in the explanation are expressed. Both informativeness and clarity are rated on a 7-point Likert scale, ranging from 1 (not informative/unclear) to 7 (very informative/very clear). We collect annotations from both trained annotators and ChatGPT to compare their assessments of explanation quality.

\subsection{Datasets}

We examine ChatGPT through the lenses of logical reasoning, misinformation justification, and implicit hate speech explanation, encompassing a spectrum of complexities and nuances of human language and cognition. 
We use three datasets: the e-SNLI  (logical reasoning)~\cite{e-snli}, the LIAR-PLUS (fake news justification)~\cite{liar_2018}, and the Latent Hatred dataset (implicit hate speech explanation)~\cite{elsherief-etal-2021-latent}.

\textbf{e-SNLI dataset.} One data instance includes a premise, a hypothesis, a label ({contradiction, neutral, or entailment}), and an explanation of the relationship between the premise and hypothesis. 
The label and explanations are annotated by experienced human annotators (native English speakers) hired on the Amazon Mechanical Turk platform~\cite{e-snli}. 
We select the e-SNLI dataset to examine the capability of LLMs to understand and evaluate logical reasoning~\cite{yu-etal-2023-alert}.

\textbf{LIAR-PLUS dataset.} Claims are labeled as {pants-fire, false, barely-true, half-true, mostly-true, or true}, and justifications for those claims, which may contain misinformation, are extracted from an existing corpus of fact-revealing human-written articles~\cite{liar_2018}. We consider the justification as NLE. 
We select the LIAR-PLUS dataset to examine the capability of LLMs to understand and discriminate counterfactual information within context.

\textbf{Latent Hatred dataset.} It includes explanations of why a text is considered to indicate implicit hate within the provided context. Experienced research assistants are hired to provide the ground truth NLE annotations in the Latent Hatred dataset~\cite{elsherief-etal-2021-latent}.
The explanations in the first two datasets rely on factual information, making them inherently less subjective than those in the Latent Hatred dataset, as deciding what qualifies as hateful can vary among individuals~\cite{gordon2022jury}.

For the analyses in this work, we randomly sample 100 data instances from each of the three datasets, keeping the label distribution similar to the original datasets. Table~\ref{tab:dataset_example} shows an example instance of each dataset.

\begin{table*}[ht]
    \centering
    \begin{tabular}{p{22mm}|p{55mm}|p{22mm}|p{40mm}}
    \toprule
    Dataset & Document \textbf{[Label]} & Auxiliary \newline Information & NLE \\
    \midrule
    e-SNLI (Logical reasoning) & 3 young man in hoods standing in the middle of a quiet street facing the camera. \textbf{[Entailment]} & Three hood wearing people pose for a picture. & People that are facing a camera are ready to pose for a picture. \\
    \midrule
    LIAR-PLUS (Misinformation justification) & A proposed tax to fund transportation projects would spend \$90,000 to take a single vehicle off the road during the morning and afternoon commute.  \textbf{[False]} & Spoke by Steve Brown during a forum hosted by The Atlanta Journal-Constitution & In sum: Brown based his calculation on a transit cost estimate thats open to accusations of cherry-picking. Furthermore, even Brown and Ross think their own number falls short. \\
    \midrule
    Latent Hatred (Implicit hate speech explanation) &  and i will point it out here when u call white people white supremacist just cause they disagree with you it s like me calling you the n word with no proof \textbf{[Implicitly hateful]} & - & The post is implicitly hateful because it implies that Blacks blame whites. \\
    \bottomrule
    
    \end{tabular}
    \caption{The data instance examples of NLE and auxiliary for the three datasets.} 
    \label{tab:dataset_example}
\end{table*} 

\subsection{Evaluation of NLE Quality}
\subsubsection{Human Annotation}  

We initially attempted to annotate the data using Amazon Mechanical Turk (MTurk). To ensure the quality of the annotations collected from MTurk, we require the workers to fulfill the following requirements: (1) to have the history approval rate equal to or more than 98\%; (2) to have the history approval instances equal to or more than 5,000; and (3) to live in the United States. However, we find that the results are not reliable enough---its inter-annotator agreement score is very low with a Krippendorff's $\alpha$ of 0.139, which have also been observed by recent studies reporting debatable quality of annotations by MTurk workers for subjective tasks~\cite{clinciu2021study}. 
Thus, we opted to hire trained research assistants for the annotation. 

We hired experienced graduate students from the institute as research assistants (RAs). To ensure they possess similar standard rating criteria as we intended, we conducted training sessions and examined them via qualification tests. 
In the training session, we provided each dataset's NLE examples of low, moderate, and high quality. 
The RAs were then required to pass the qualification test. 
The qualification test comprised five NLEs asking for the informativeness and clarity scores separately, requiring them to judge the coarse-grained explanation quality (selecting from {low}, {moderate}, and {high} levels of quality) of given NLEs with their auxiliary information attached. The ground truth labels of those questions were selected from the data annotated by the experienced MTurk workers and then verified by three separate human experts. 
Only those who provide the correct answers to at least nine out of ten questions are invited for further annotations.  
If an RA fails this initial assessment, we provide another training session and have the second qualification test.
Two of the eight candidate RAs failed the first qualification test. One of those two who did not pass the first test also failed the second test. As a result, we hired seven RAs (four females and three males) as our annotators. 

The RAs evaluate the informativeness and clarity of randomly selected NLP instances. Three different annotators assess each of the 300 instances. 
The process takes 30 working hours, with each annotator careful not to evaluate the same instance twice. They are compensated a total of \$450 based on their working hours.
During annotation, RAs are instructed to assign scores to each instance within a reasonable time frame. We recommend allocating approximately 60 seconds to rating each instance. However, if they encounter instances that are long or challenging to assess, they are encouraged to take additional time to rate.

The inter-coder reliability scores for the e-SNLI, LIAR-PLUS, and Latent Hatred datasets are 0.7416, 0.6736, and 0.6755, respectively. As expected, the e-SNLI dataset, which focuses on logical reasoning statements, achieves the highest agreement score among raters. Conversely, the Latent Hatred dataset receives a lower score, suggesting that raters' differing backgrounds may lead to varied interpretations of the same text. Surprisingly, the LIAR-PLUS dataset shows an even lower agreement score, indicating that discerning the justifications linked to claims is a more challenging task than expected. This analysis demonstrates that the datasets selected for evaluation require different levels of understanding and reasoning skills, as evidenced by the inter-coder reliability scores.

Table~\ref{tab:annotation_compare} summarizes the inter-annotator agreement scores, elapsed time, the average annotation speed per data instance, and the total expenses. 
The inter-annotator agreement was high with an $\alpha$ of 0.721, considered acceptable agreement~\cite{clinciu2021study}. 
We use these annotations as ground truth to compare with ChatGPT's annotations.

\begin{table*}[ht]
    \centering
    \begin{tabular}{p{25mm}|p{23mm}|p{30mm}|p{28mm}|p{20mm}}
    \toprule
    Annotator Source & Krippendorff's $\alpha$ & Time Consumption & Annotation Speed per Instance & Expenses \\
    \midrule
    Trained RAs & {0.721} &  30 hours & 120 seconds & \$450 \\
    \midrule
    ChatGPT & 0.743 & 2 hours & 3.5 seconds & \$2 \\
    \bottomrule
    \end{tabular}
    \caption{Characteristics of annotations by trained RAs and ChatGPT.}
    \label{tab:annotation_compare}
\end{table*}

\subsubsection{ChatGPT Annotation}

We query the ChatGPT ({gpt-3.5-turbo}  released on May 24, 2023)~\cite{openai_chatgpt_version} to assess the informativeness and clarity of each NLE for the 300 sampled instances. For each instance, we query three times and take the average scores using its official API provided by OpenAI~\cite{openai_chatgpt_api}.
The prompt has two components: task description and instructions. The task description is \textit{Task: Scoring the quality of the given natural language explanation based on its referring tweet from the perspective of informativeness and clarity.} Then, the instructions are in six parts: (1) dataset description, (2) metric description (what informativeness and clarity are), (3) high score examples, (4) low score examples, (5) input (e.g., tweet, annotated label, and explanation), and (6) instruction (asking for the two ratings). 
Each generation is from a distinct session to rule out the effects of chat context.
The ChatGPT annotation took two hours and cost \$2 in total. 
While coming from the same model, the answers are relatively consistent with the $\alpha$ of 0.743.

\section{RQ1: Does ChatGPT's evaluation of the quality of NLEs align with human assessments?}
\label{sec:rq1}

Withholding the annotated data at hand, in this section, we attempt to evaluate how accurately ChatGPT can assess the explanation quality of NLEs. 
In doing so, we first analyze the correlations between the two evaluation results. Then, we formulate it as a classification task: using the human (i.e., trained RAs) evaluations of the quality of explanations as ground truth data, we use ChatGPT evaluations as predictions and evaluate its performance in classifying the level of the quality of explanations. 

\subsection{Correlation Analysis}

Using correlation analysis to explain the relationship between ChatGPT-generated ratings and ground-truth human ratings is one practical protocol to present the capability of the language model in providing subjective ratings of NLEs.
By statistically analyzing the correlation between ChatGPT-generated ratings and RAs annotations, we can delineate the scenarios where the model acts perfectly and cases where it might require further tuning and adjustments.

\begin{figure*}[!tbp]
  \centering
  
  \begin{subfigure}[b]{0.32\textwidth}
    \centering
    \includegraphics[width=\textwidth]{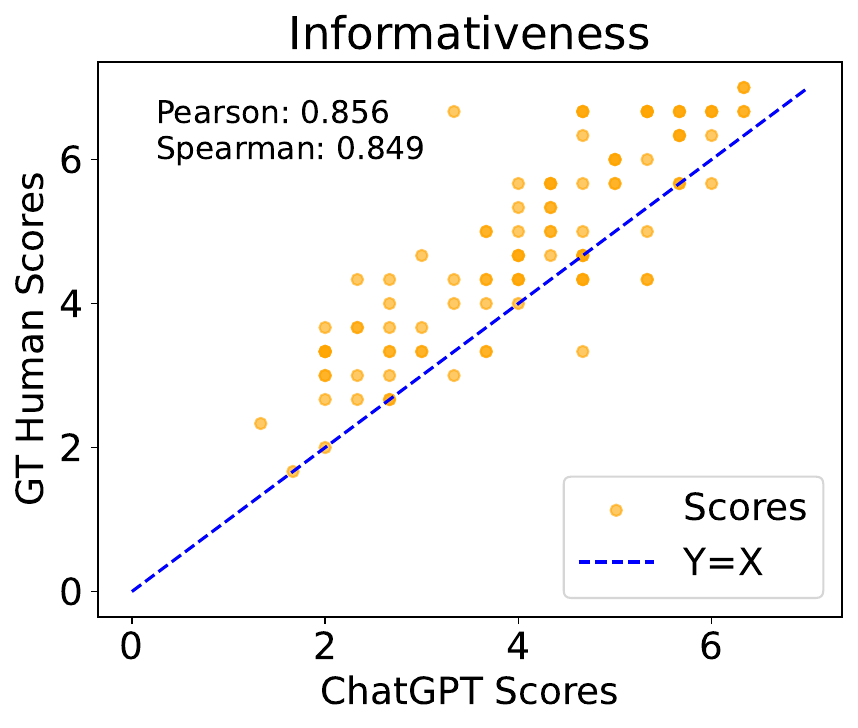}
  \end{subfigure}
  \hfill
  \begin{subfigure}[b]{0.32\textwidth}
    \centering
    \includegraphics[width=\textwidth]{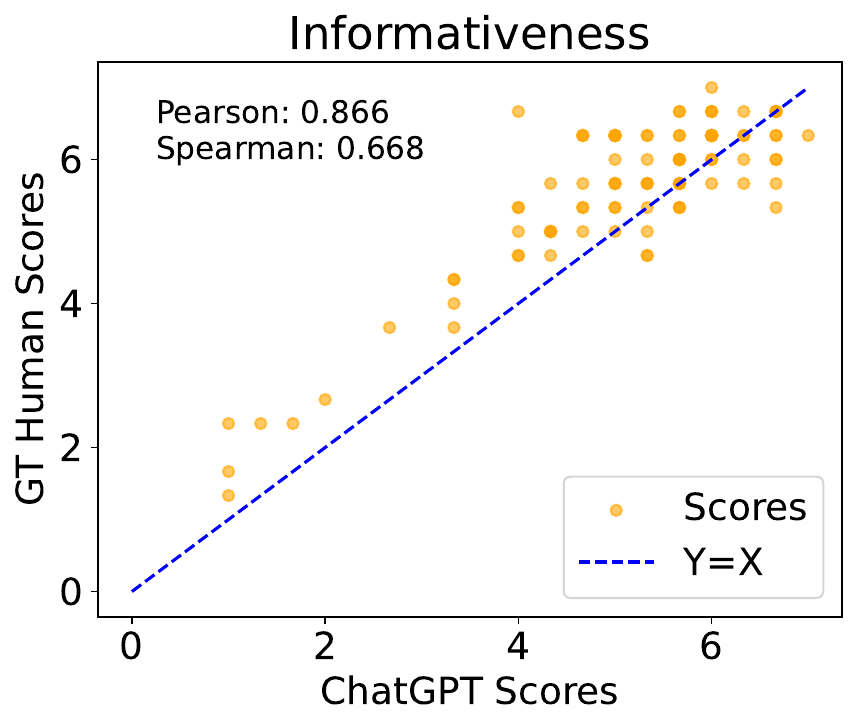}
  \end{subfigure}
  \hfill
  \begin{subfigure}[b]{0.32\textwidth}
    \centering
    \includegraphics[width=\textwidth]{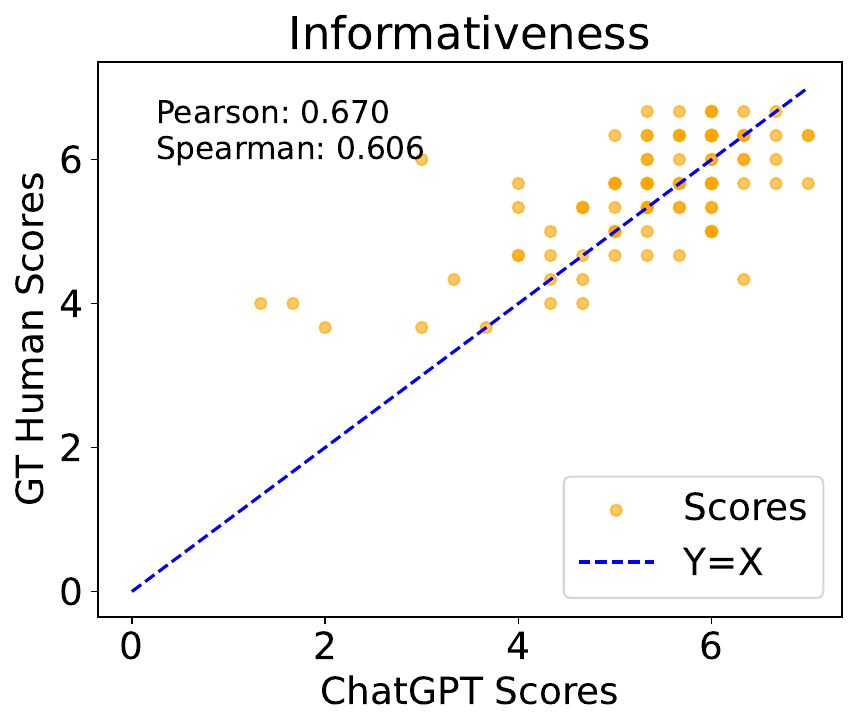}
  \end{subfigure}

  \begin{subfigure}[b]{0.32\textwidth}
    \centering
    \includegraphics[width=\textwidth]{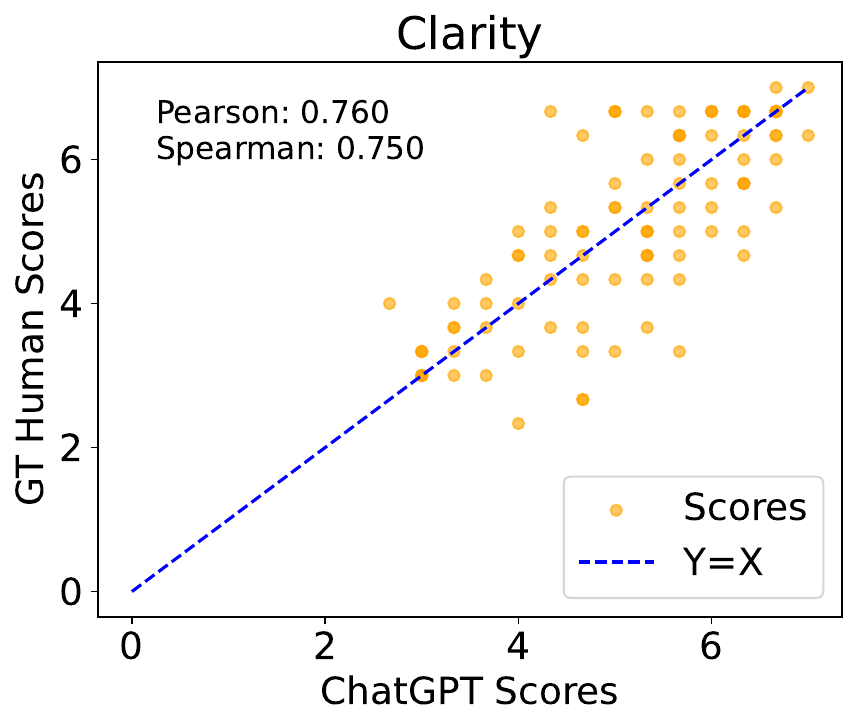}
    \caption{Logical Reasoning}
  \end{subfigure}
  \hfill
  \begin{subfigure}[b]{0.32\textwidth}
    \centering
    \includegraphics[width=\textwidth]{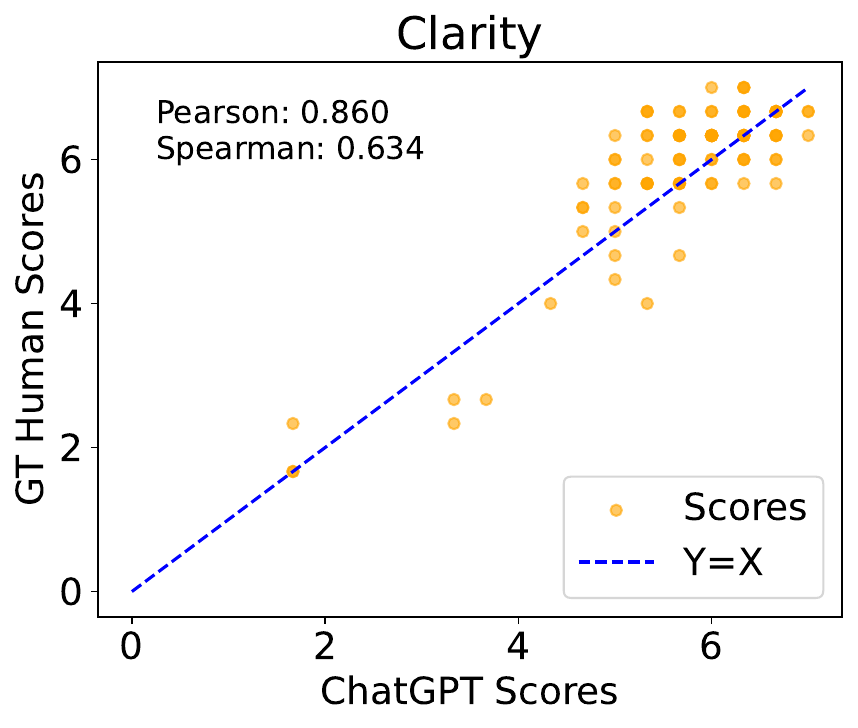}
    \caption{Misinformation Justification}
  \end{subfigure}
  \hfill
  \begin{subfigure}[b]{0.32\textwidth}
    \centering
    \includegraphics[width=\textwidth]{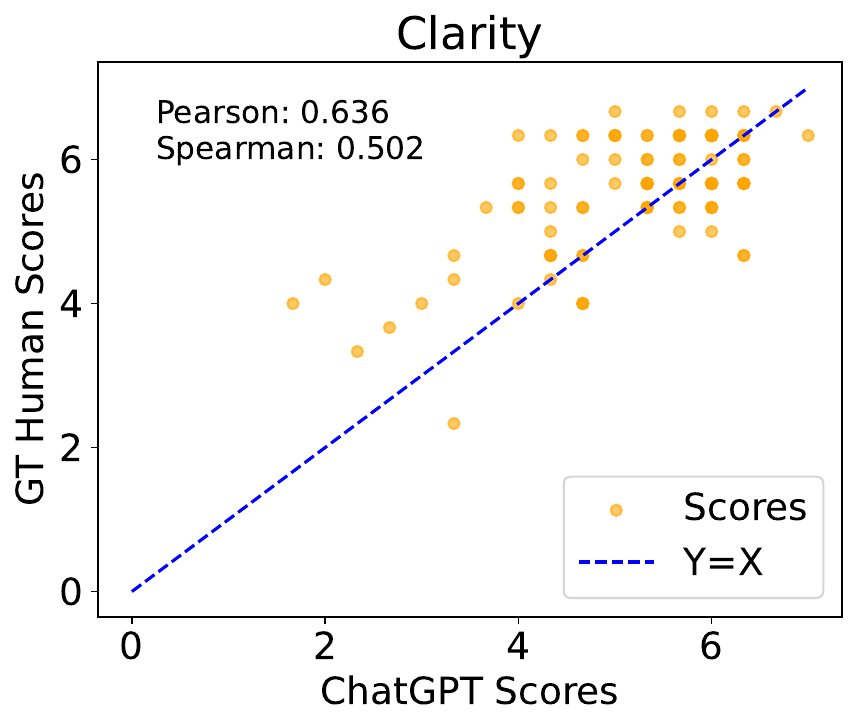}
    \caption{Implicit Hate Speech Explanation}
  \end{subfigure}
  \caption{Correlation between ChatGPT and human evaluations for the three datasets and the two metrics, Informativeness and Clarity. Pearson's and Spearman's correlation coefficients are in the figure.}  
  \label{fig: correlation}
\end{figure*}

Figure~\ref{fig: correlation} visualizes the correlations between human and ChatGPT ratings where the blue dashed line indicates a $y=x$ relationship. 
We can observe strong positive correlations between the evaluations for informativeness and clarity across all three datasets.
Pearson's correlation coefficients (Spearman's ranking correlation coefficients) range from 0.636 (0.502) to 0.888 (0.849). 
In most cases, ChatGPT tends to underestimate the quality, especially for the informativeness of all three datasets, except for the clarity on logical reasoning and the informativeness scores on implicit hate speech explanation. 
We also noticed that the clarity ratings show weaker correlations than the informativeness ratings, which might be due to the inherent ambiguity of the definition of clarity. 
ChatGPT ratings correlate well with human ratings for logical reasoning and misinformation justification datasets, while the correlation for the implicit hate speech explanation is weaker. 
This is unsurprising, as the implicit hate speech explanation dataset's explanations tend to be shorter than the other two datasets. Consequently, a significant portion of their scores falls within the range of 4 to 7, making it challenging to distinguish them.

\subsection{Coarse- and Fine-Grained Assessment via Classification}

\begin{table}[ht]
    \centering
    \begin{tabularx}{0.48\textwidth}{c|cll}
    \toprule
    Dataset & Metric & F1-score$\uparrow$ & RMSE$\downarrow$ \\
    \midrule
    e-SNLI &  Info. & 0.88 ($\textbf{+0.28}$) & 0.36 ($-0.17$) \\
             &  Clar. & 0.85 ($+0.18$) & 0.37 ($-0.11$) \\
    \midrule
    LIAR &  Info. & 0.97 ($+0.09$) & 0.17 ($-0.11$) \\
    -PLUS  &  Clar. & \textbf{1.00} ($+0.09$) & \textbf{0.00} ($\textbf{-0.24}$) \\
    \midrule
    Latent &  Info. & 0.97 ($+0.01$) & 0.20 ($+0.03$)\\
    Hatred &  Clar. & 0.95 ($-0.01$) & 0.24 ($+0.08$)\\
    \bottomrule
    \end{tabularx}
    \caption{Binary (low and high) classification results for the two metrics across the three datasets. We report the weighted f1-score and RMSE (Root Mean Square Error). The numbers in parentheses show the improvement from the baseline.}
    \label{tab:G2}
\end{table}

\begin{table}[ht]
    \centering
    \begin{tabularx}{0.48\textwidth}{c|cll}
    \toprule
    Dataset & Metric & F1-score$\uparrow$ & RMSE$\downarrow$ \\
    \midrule
    e-SNLI &  Info. & 0.64 ($+0.34$) & 0.62 ($\textbf{-0.24}$)\\
             &  Clar. & 0.77 ($\textbf{+0.34}$)& 0.47 ($\textbf{-0.24}$)\\
    \midrule
    LIAR &  Info. & 0.84 ($+0.07$)& 0.44 ($-0.14$)\\
    -PLUS & Clar.  & \textbf{0.90} ($+0.06$)& \textbf{0.33} ($-0.21$)\\
    \midrule
    Latent &  Info. & 0.87 ($+0.13$)& 0.37 ($-0.05$)\\
    Hatred &  Clar. & 0.81 ($+0.10$)& 0.47 ($-0.01$)\\
    \bottomrule
    \end{tabularx}
    \caption{Ternary (low, moderate, and high) classification results.}
    \label{tab:G3}
\end{table}

\begin{table}[ht]
    \centering
    \begin{tabularx}{0.48\textwidth}{c|cll}
    \toprule
    Dataset & Metric & F1-score$\uparrow$ & RMSE$\downarrow$ \\
    \midrule
    e-SNLI &  Info. &  0.36 ($+0.26$)& 0.99 ($-0.40$)\\
             &  Clar. &   0.51 ($\textbf{+0.35}$)& 0.89 ($\textbf{-0.92}$)\\
    \midrule
    LIAR &  Info. & 0.53 ($+0.21$)& 0.79 ($-0.62$)\\
    -PLUS &  Clar. & \textbf{0.58} ($+0.08$)& \textbf{0.66} ($-0.60$)\\
    \midrule
    Latent &  Info. & 0.58 ($+0.30$)& 0.86 ($+0.06$)\\
    Hatred &  Clar. & 0.46 ($+0.23$)& 0.95 ($+0.09$)\\
    \bottomrule
    \end{tabularx}
    \caption{7-way (integer bins from 1 to 7) classification results.}
    \label{tab:G6}
\end{table}

To quantify the capability of ChatGPT in evaluating the quality of NLEs, we formulate the classification tasks on three granularities, from the binary classification of distinguishing low/high to the ternary classification of low/moderate/high, and the raw scores (1 to 7). 
We employ the weighted F1-score and root mean square error (RMSE) as evaluation metrics in presenting our results. While the f1-score is a common measure for classification problems, they may overlook errors arising from variations in category orders. RMSE, on the other hand, accounts for these order differences.

We begin with coarse granularity. We test whether ChatGPT can assess the explanation quality of NLEs in a \emph{binary} way, i.e., high and low. 
We categorize informativeness and clarity scores falling between 1 and 4 as low quality and scores between 4 and 7, including 4, as high quality.
Table~\ref{tab:G2} presents the binary classification result and the performance changes compared to the baseline models. We consider the majority voting our baseline model (i.e., label all the instances using the most prominent label). We expect to observe high f1-scores and low RMSE in binary classifications as the data is imbalanced (most data points are gathered in the `high' class).
The ChatGPT performs very well, especially with 100$\%$ correctness in measuring clarity on a binary scale for the misinformation justification dataset.
ChatGPT performs best for misinformation justification and worst for logical reasoning. As for the performance improvements, the biggest is for the logical reasoning dataset, and the least is for the implicit hate speech explanation dataset.

Next, we explore ChatGPT's performance in more detail. To address this, we establish a ternary classification task, with the classes defined as follows: low (ranging from 1 to 3), moderate (spanning from 3 to 5), and high (encompassing the range from 5 to 7). Table~\ref{tab:G3} shows the results of this ternary classification. Similarly, the ChatGPT performs the best for the misinformation justification dataset and the worst on the logical reasoning dataset to provide quality rating scores. Compared with the baseline, the ChatGPT provides the biggest performance improvements for the logical reasoning dataset while the least for the implicit hate speech explanation dataset.

\begin{figure*}[!tbp]
  \centering
  
  \begin{subfigure}[b]{0.32\textwidth}
    \centering
    \includegraphics[width=\textwidth]{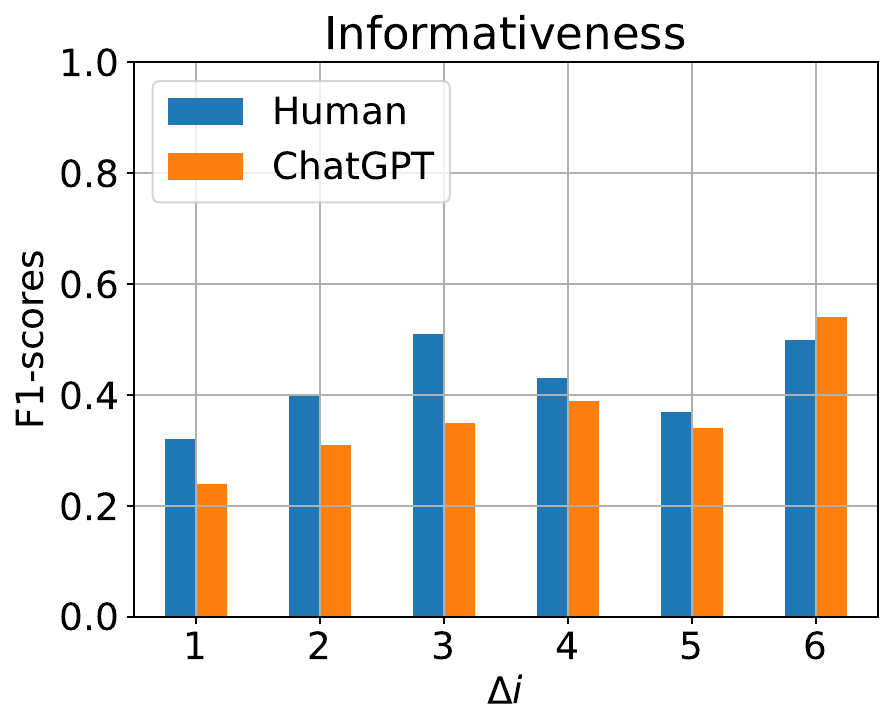}
  \end{subfigure}
  \hfill
  \begin{subfigure}[b]{0.32\textwidth}
    \centering
    \includegraphics[width=\textwidth]{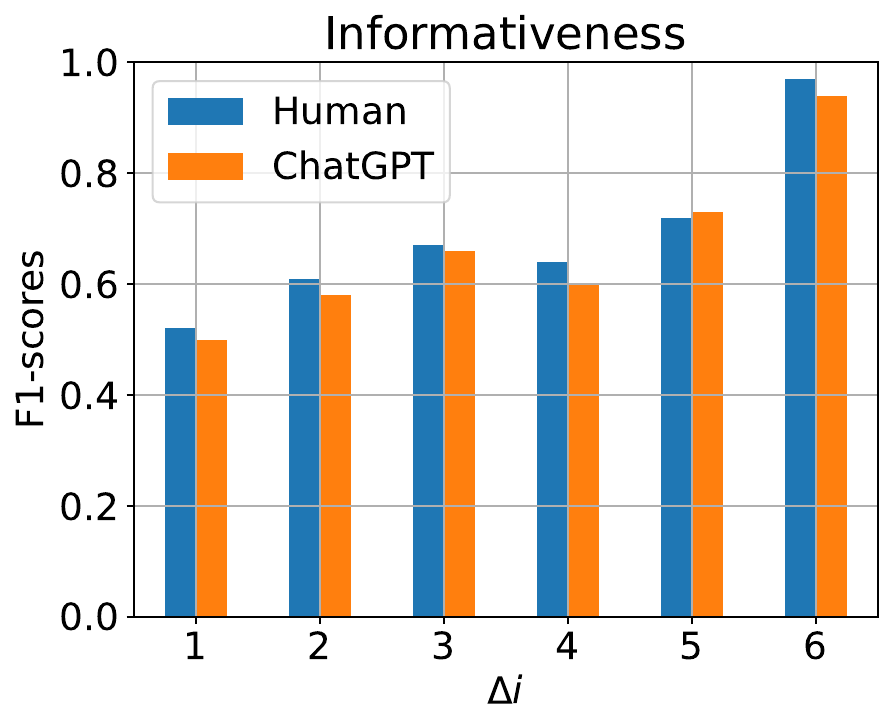}
  \end{subfigure}
  \hfill
  \begin{subfigure}[b]{0.32\textwidth}
    \centering
    \includegraphics[width=\textwidth]{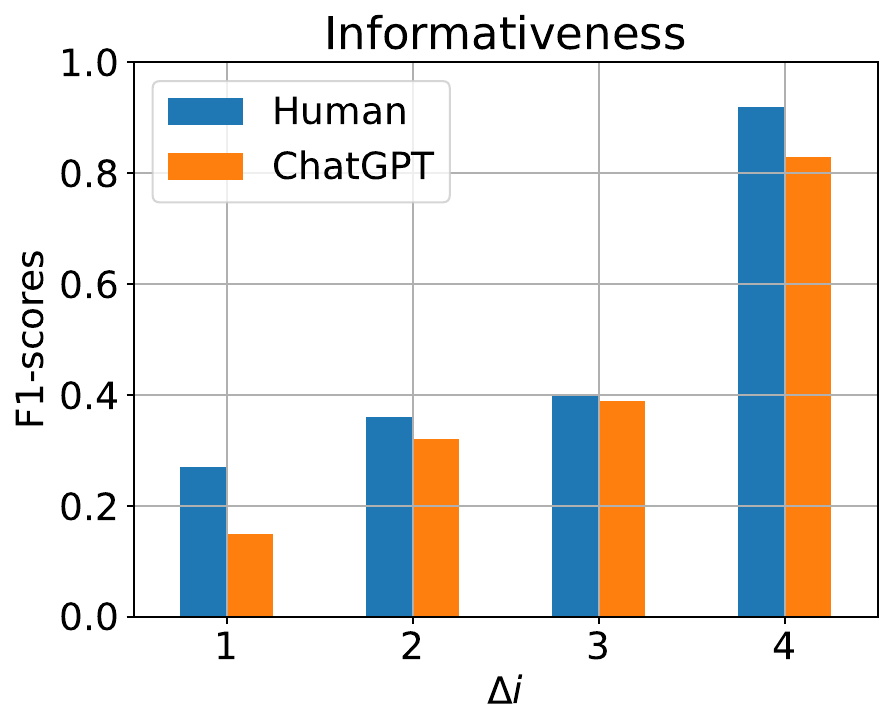}
  \end{subfigure}

  \begin{subfigure}[b]{0.32\textwidth}
    \centering
    \includegraphics[width=\textwidth]{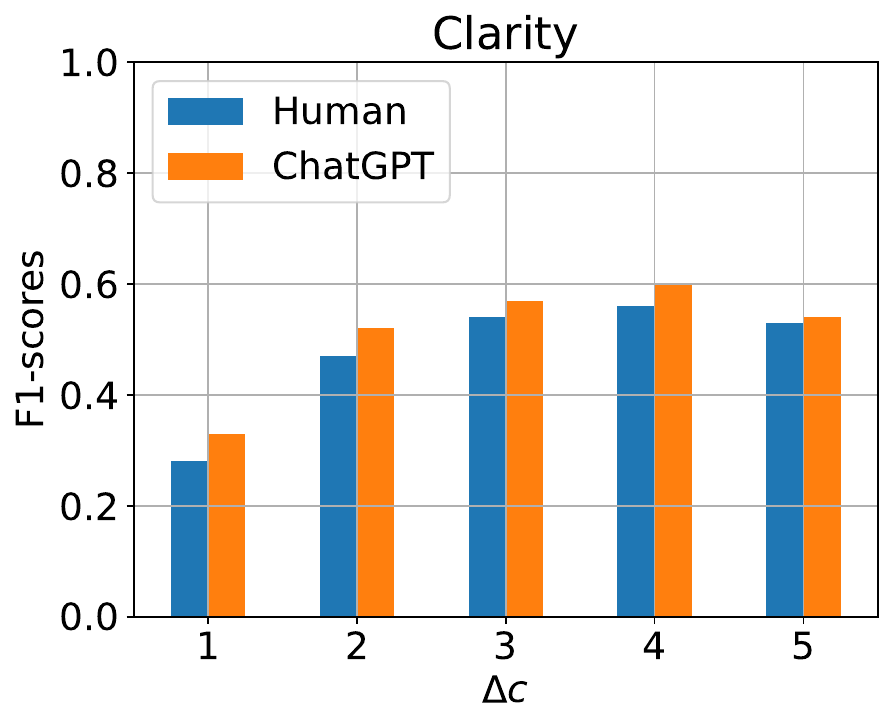}
    \caption{Logical Reasoning}
  \end{subfigure}
  \hfill
  \begin{subfigure}[b]{0.32\textwidth}
    \centering
    \includegraphics[width=\textwidth]{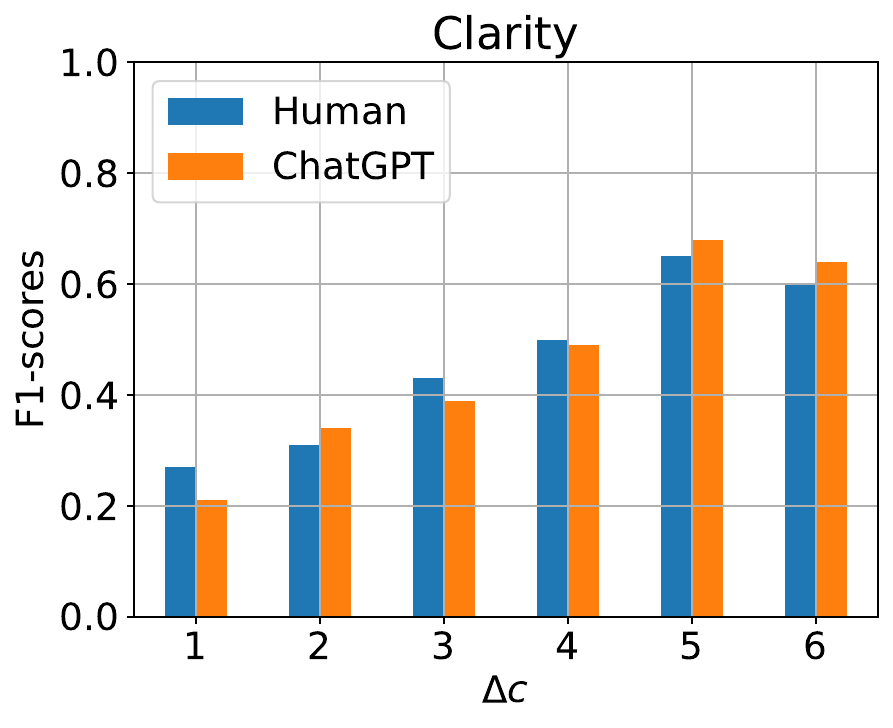}
    \caption{Misinformation Justification}
  \end{subfigure}
  \hfill
  \begin{subfigure}[b]{0.32\textwidth}
    \centering
    \includegraphics[width=\textwidth]{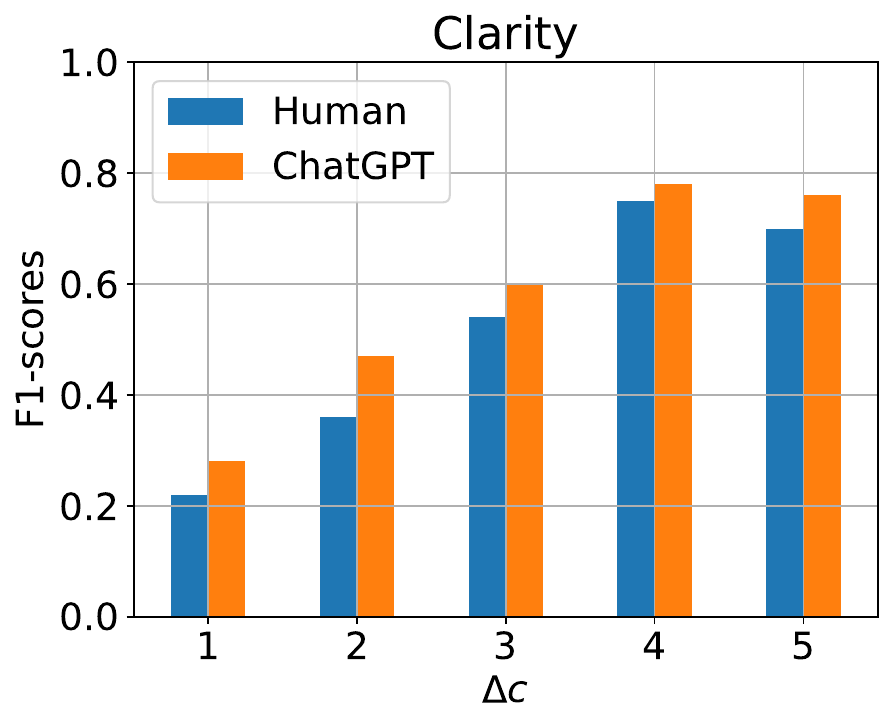}
    \caption{Implicit Hate Speech Explanation}
  \end{subfigure}
  
  \caption{Visualization of ChatGPT pair comparison f1-scores for various $\Delta_{i}$ and $\Delta_{c}$, from 0 to 6 rounded by integer bins, compared with additional human annotations specifically for the pair comparison task.}
  \label{fig: pair-comparison}
\end{figure*}

We also use the raw informativeness and clarity score for the finest granularity of score estimation as in Table~\ref{tab:G6}. We can observe similar patterns where the ChatGPT performs the best for the misinformation justification dataset. 
The ChatGPT performs the best for the clarity rating scores (0.58 f1-score and 0.66 RMSE value) while the worst for informativeness rating scores (0.36 f1-score and 0.99 RMSE value). As for performance improvement, ChatGPT performs best regarding clarity rating scores (+0.35 f1-score and -0.92 RMSE value). 
The performance disparities may arise from LLMs having different domain knowledge, as observed in existing works~\cite{NEURIPS2021_inductive_bias, openai_gpt4v, yang2023dawn, huang2024token}.

In summary, among all three granularities, we observe a strong alignment between ChatGPT evaluation and human assessment and similar patterns of the best performance for misinformation justification rating and the worst for logical reasoning rating. However, as the granularity increases, the best f1-score drops from 1.0 to 0.58 (clarity for misinformation justification dataset), which means that the ChatGPT only performs well in the coarse-grained rating.

\section{RQ2: Is ChatGPT capable of comparing two NLEs in terms of their explanation quality?}
\label{sec:rq2}

We evaluate ChatGPT's capability via pairwise comparison. 
Comparing instances is easier and more accurate than giving a score to an instance for human annotators~\cite{NIPS2006_trueskill}. 
Our results in $\S$\ref{sec:rq1} also show that fine-grained quality assessment is challenging. 
If ChatGPT excels in performing pairwise comparisons close to human judgment, it becomes feasible to obtain precise estimations for all instances using pairwise comparison results~\cite{NIPS2006_trueskill}.

In this experiment, rather than requesting ratings for explanation quality, we present pairs of two explanations and ask ChatGPT and human annotators which one is perceived as more informative or clear. 
Those pairs with small score differences are more challenging to discern than those with larger ones. We analyze the performance by varying score differences $\Delta$. 
We denote the score differences of the ground-truth dataset as $\Delta_{i}$ for informativeness and $\Delta_{c}$ for clarity. 
Then, for each of $\Delta_{i}$ and $\Delta_{c}$, we randomly select 100 instances out of all possible pairs.  
We use all pairs for cases where it does not have 100 pairs (e.g., implicit hate speech explanation dataset only has 41 pairs at most when $\Delta_{i} = 4$).
In total, we use 520 (520), 550 (546), and 321 (372) pairs for informativeness (clarify) for logical reasoning, misinformation justifications, and implicit hate speech explanation datasets, respectively.
We collected 8,346 annotations from both human annotators and ChatGPT. The same trained RAs provided the annotations and were compensated a total of \$650 based on their working hours for this task. 
We classify the outcomes of the pairwise comparisons into three categories: 1) the first instance has a higher score, 2) the first instance has an equal score to the second, and 3) the first instance has a lower score.

\begin{figure*}[!tbp]
  \centering
  
  \begin{subfigure}[b]{0.32\textwidth}
    \centering
    \includegraphics[width=\textwidth]{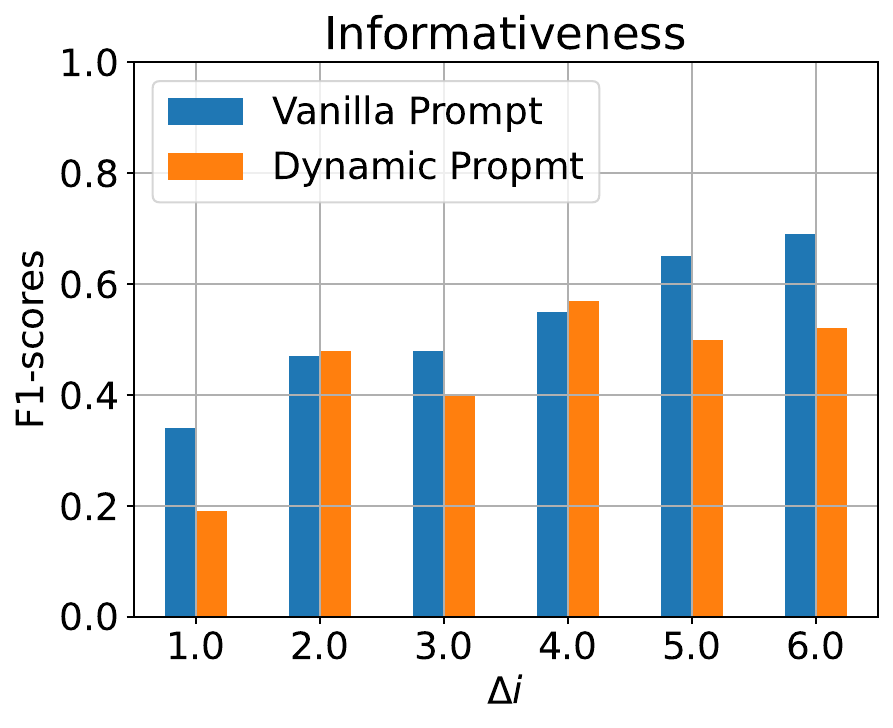}
  \end{subfigure}
  \hfill
  \begin{subfigure}[b]{0.32\textwidth}
    \centering
    \includegraphics[width=\textwidth]{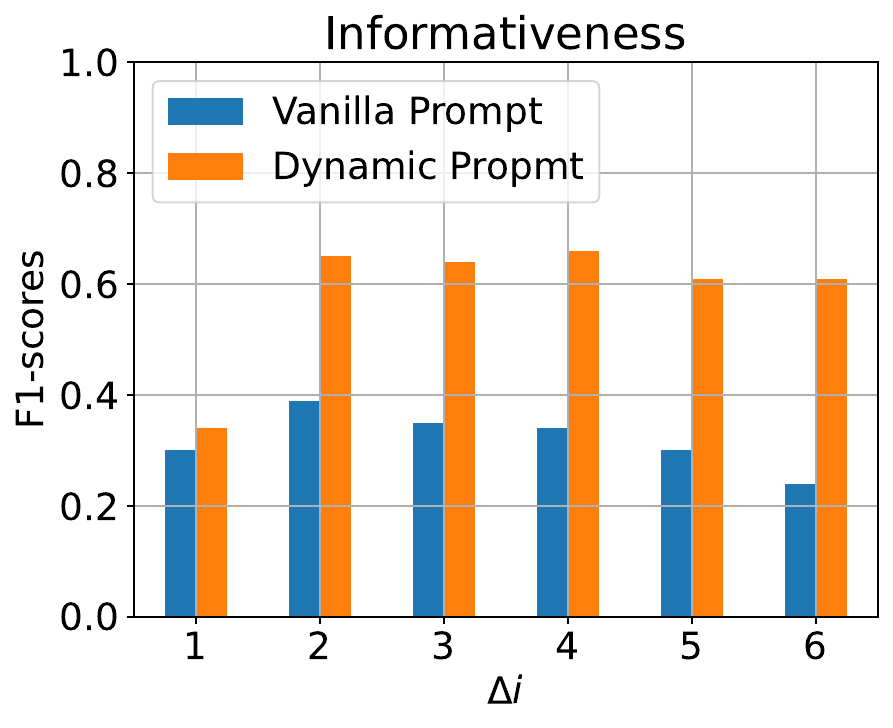}
  \end{subfigure}
  \hfill
  \begin{subfigure}[b]{0.32\textwidth}
    \centering
    \includegraphics[width=\textwidth]{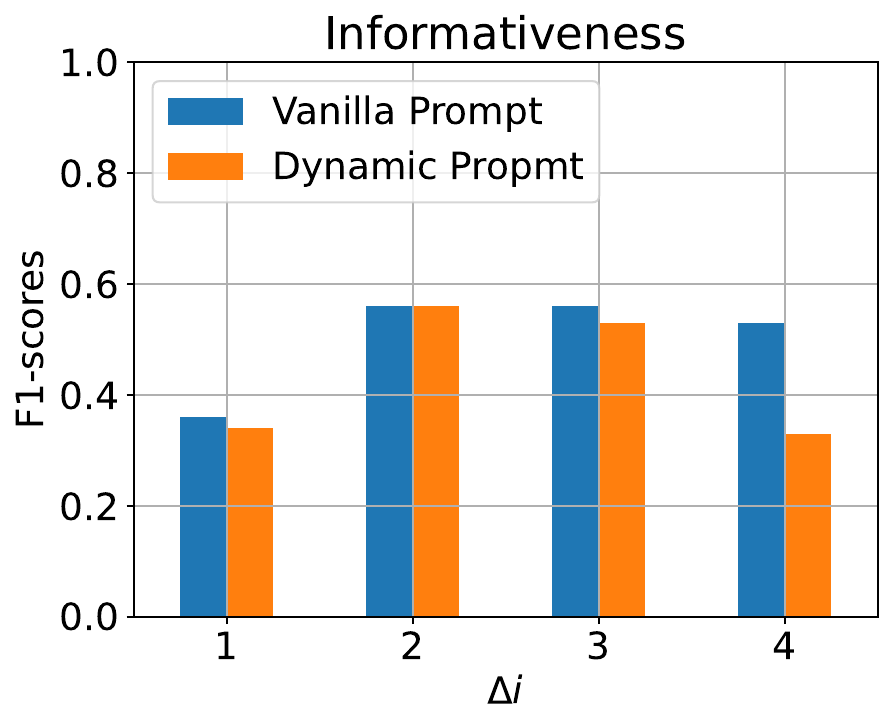}
  \end{subfigure}

  \begin{subfigure}[b]{0.32\textwidth}
    \centering
    \includegraphics[width=\textwidth]{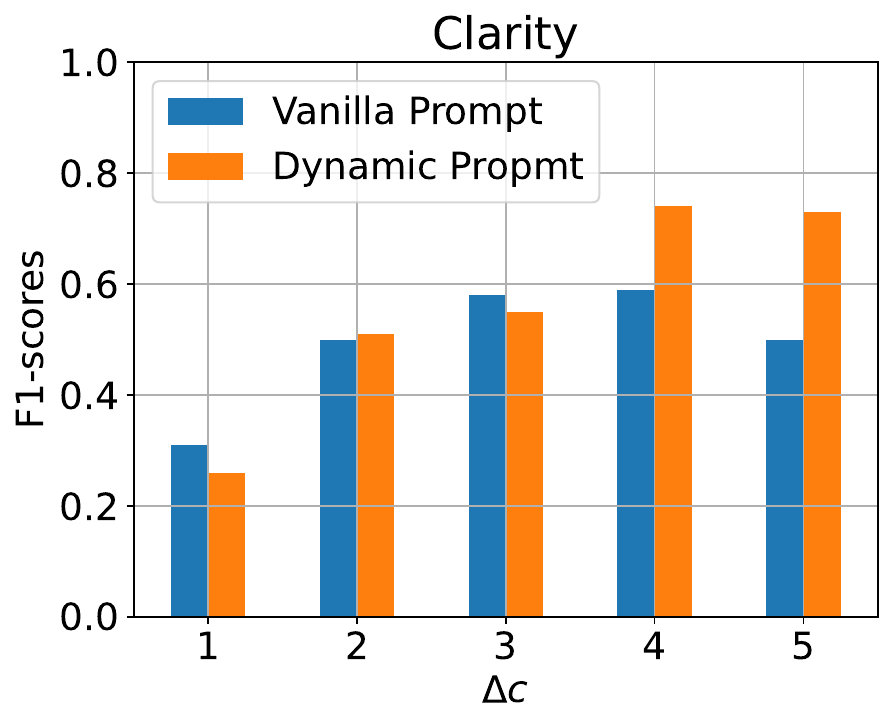}
    \caption{Logical Reasoning}
  \end{subfigure}
  \hfill
  \begin{subfigure}[b]{0.32\textwidth}
    \centering
    \includegraphics[width=\textwidth]{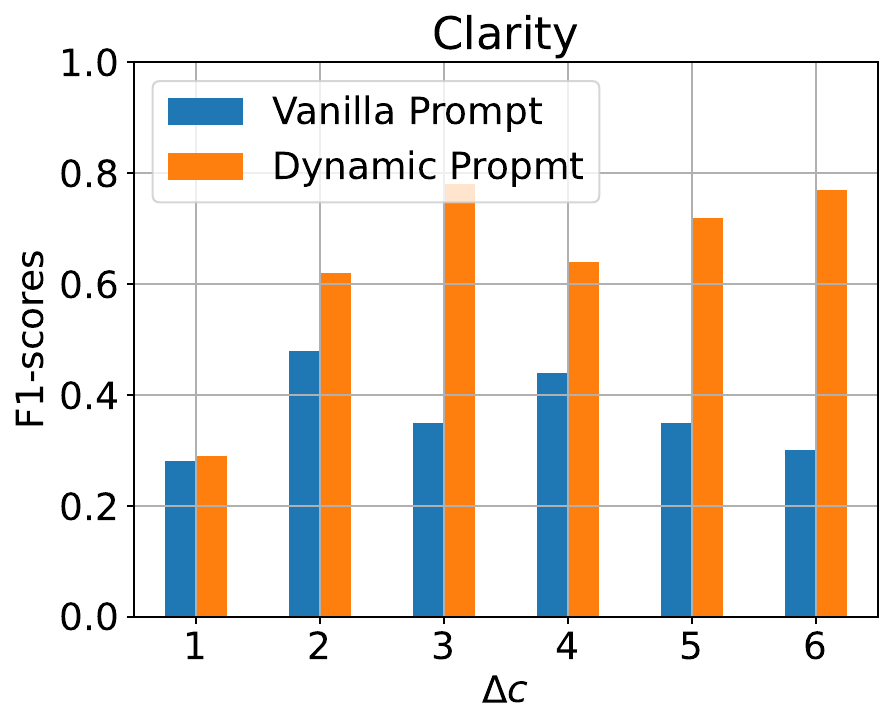}
    \caption{Misinformation Justification}
  \end{subfigure}
  \hfill
  \begin{subfigure}[b]{0.32\textwidth}
    \centering
    \includegraphics[width=\textwidth]{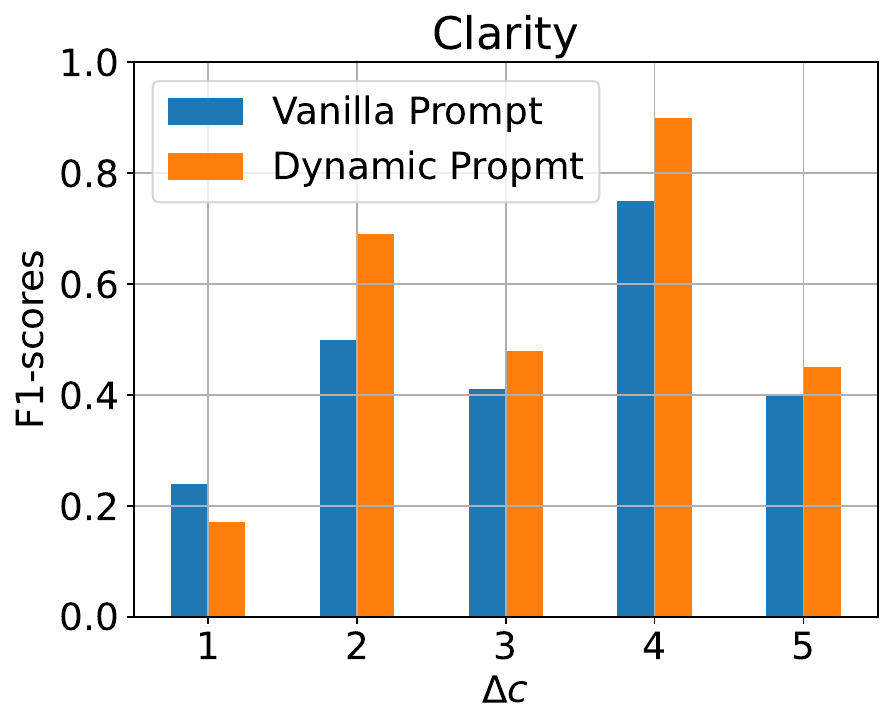}
    \caption{Implicit Hate Speech Explanation}
  \end{subfigure}
  \caption{Visualization of ChatGPT pair comparison f1-scores for various $\Delta_{i}$ and $\Delta_{c}$, from 0 to 6 rounded by integer bins, showing the difference between vanilla prompting and our proposed dynamic prompting. The measurement is only based on NLE evaluation metrics of Informativeness and Clarity.}
  \label{fig: pair-comparison-DP}
\end{figure*}

Figure~\ref{fig: pair-comparison} shows how accurate the pairwise comparison is by varying $\Delta_{i}$ and $\Delta_{c}$. As $\Delta_{i}$ and $\Delta_{c}$ increase, ChatGPT and human experts generally perform better to distinguish the difference from NLE pairs. 
For the sixteen $\Delta_{i}$ cases, human outperforms ChatGPT by 4.63\%p of f1-score; for the sixteen $\Delta_{c}$ cases, ChatGPT outperforms human by 3.06\%p f1-score. 
We observed a similar pattern for all three datasets: ChatGPT performs better than human experts in clarity assessment and worse in informativeness assessment. Still, for the misinformation justification dataset, the clarity assessment performs poorly when the $\Delta_{c}$ is one, three, or four. 
Human experts and ChatGPT can distinguish well for the implicit hate speech explanation when the $\Delta_{i}$ and $\Delta_{c}$ increase. Specifically for the informativeness ratings of logical reasoning and implicit hate speech explanation datasets, when $\Delta_{i}$ is small, ChatGPT performs poorly, and even when $\Delta_{i}$ becomes 3, the overall performance is still less than 0.40 f1-score.

In summary, ChatGPT performs worse than human annotations for most informativeness comparison cases (average difference of 4.63\%p) but mostly slightly better for clarity comparison cases (average difference of 3.06\%p). 
ChatGPT performs the best for explanation quality comparison of the misinformation justification dataset among three datasets. Still, it performs significantly worse than human annotators when distinguishing a small difference in rating scores (i.e., informativeness rating of logical reasoning and implicit hate speech explanation datasets).
Considering the average absolute f1-score difference among the three datasets, ChatGPT performs better in misinformation justification (2.92\%p)  than logical reasoning (5.64\%p) and implicit hate speech explanation (5.80\%p) datasets. In the next section, we will explore how to improve ChatGPT's performance on quality evaluation further.

\section{RQ3: Can dynamic prompting enhance ChatGPT's ability to assess NLE  quality?}

\begin{table}[ht]
    \centering
    \begin{tabularx}{0.48\textwidth}{c|cll}
    \toprule
    Dataset & Metric & F1-score$\uparrow$ & RMSE$\downarrow$ \\
    \midrule
    e-SNLI &  Info. & 0.25 ($-0.09$)& 1.88 ($+0.45$)\\
             &  Clar. & 0.19 ($0.00$)& 2.12 ($+0.51$)\\
    \midrule
    LIAR &  Info. & 0.38 ($-0.02$)& \textbf{0.99} ($+0.11$)\\
    -PLUS         &  Clar. & \textbf{0.46} ($\textbf{+0.11}$)& 1.06 ($\textbf{+0.01}$)\\
    \midrule
    Latent &  Info. & 0.19 ($-0.17$)& 1.35 ($+0.07$)\\
    Hatred &  Clar. & 0.32 ($+0.05$)& 1.91 ($+0.61$)\\
    \bottomrule
    \end{tabularx}
    \caption{Dynamic prompting results.  The granularity used here is 7-way classification.} 
    \label{tab:G6-DP}
\end{table}

As in-context learning helps LLMs to understand a task~\cite{khalifa2022lepus} better, we explore dynamic prompting for fine-grained assessment and pair-comparison tasks. 
The idea is to provide a prompt with the most semantically similar example instead of fixed examples, along with its informativeness and clarity scores.
To identify the most similar example, we first split our data into a candidate example set and a test set, with the specific division method detailed later as it varies across experiments. Then, we obtain the representations of every explanation using SentenceBERT~\cite{reimers2019sentence}. By using a pre-trained model\footnote{The paraphrase-distilroberta-base-v1 in sentence-transformers:~\url{https://github.com/UKPLab/sentence-transformers}.}, we choose the one with the highest cosine similarity from the candidate example set for each instance in the test set.

We evaluate the performance of dynamic prompting by fine-grained classification and pairwise comparison.
We use the same ground-truth datasets built in RQ1 and RQ2. 
We employ a 50:50 data split for each dataset, with one half as the candidate explanation examples and the other half as the test set.
In the case of the pairwise comparison, we utilize a 50:50 data split for each $\Delta$ within each dataset, each containing a maximum of 100 pairs.
We compare the performance of dynamic prompting with 
that of the prompting used in $\S$\ref{sec:rq1} and $\S$\ref{sec:rq2}. We note that the baseline results differ from those presented in Figure~\ref{fig: pair-comparison} as we use 50\% of the data for testing in this analysis.

As shown in Table~\ref{tab:G6-DP}, the dynamic prompting does not improve classification performance but worsens it. Dynamic prompting works relatively equivalent to the vanilla prompt design only for the misinformation justification dataset. Generally, the clarity metric performance becomes slightly better with in-context learning, and the informativeness metric performance significantly worsens. 
We could not find any evidence that dynamic prompting improves the performance of explanation quality ratings. 

However, in the pairwise comparison, dynamic prompting shows sizable improvement for the misinformation justification dataset (Figure~\ref{fig: pair-comparison-DP}). 
On average, it provides 5.13\%p and 14.50\%p improvements of informativeness and clarity across the different $\Delta$ among three datasets.
Dynamic prompting makes it even worse for the case that showed the worst performance in the previous section: informativeness for logical reasoning and implicit hate speech explanation datasets. 

In summary, dynamic prompting can further improve the performance of ChatGPT to estimate the explanation quality for the misinformation justification dataset in a comparison setting. However, the performance for the other two datasets does not always increase through dynamic prompting under different $\Delta_{i}$ and $\Delta_{c}$. 
In a pairwise comparison setting, we notice a substantial performance improvement attributed to dynamic prompting. However, its effect is adverse in raw score classification.
This result implies that ChatGPT can better leverage the additional contextual information for comparison than directly estimate the score, especially for the misinformation justification dataset.

\section{Limitations}

Our study sheds light on the alignment between ChatGPT and human assessments in different levels of logical reasoning and implication understanding, evaluated at the subjective quality assessment of natural language explanations. However, several limitations remain for further discussion and consideration.

This work primarily evaluates ChatGPT (i.e., gpt-3.5-turbo model), which is a closed-source language model. This raises concerns about the generalizability of our findings to other LLMs, particularly open-source models. Future research should explore a wider range of LLMs to understand their capabilities in assessing NLE quality and to determine the extent to which our observations regarding ChatGPT are generalizable.

We observe ChatGPT's remarkable capability in coarse-grained categorization tasks, particularly within misinformation justification datasets. Our investigation uses a simple and replicable prompt design to facilitate research reproducibility. However, it is important to acknowledge that different prompt designs may influence ChatGPT's performance. Also, for the dynamic prompting, we only add the two examples in the prompt without altering any other information. 
Further investigation into how changes in instructions and dynamic examples within the prompt influence ChatGPT's responses would be beneficial. This exploration will help understand the impact of different prompting designs on evaluating NLE quality, focusing on the alignment between human and machine intelligence

Though robust, the metrics used in our study, Informativeness (How helpful the NLE is while understanding the context with the given labels) and Clarity (How clearly the ideas in the NLE are expressed), might not fully capture the multifaceted nature of human language and explanation quality. Future research could incorporate diverse text quality or helpfulness evaluation metrics tailored for different context-understanding tasks, enhancing our understanding of human-machine intelligence alignments.

\section{Conclusion}
ChatGPT exhibited remarkable capability in coarse-grained categorization tasks, particularly within misinformation justification datasets, while simultaneously revealing potential pitfalls, especially in finer granularity assessments in various settings. 
While robust, the metrics we applied in this study may fail to encapsulate the multilayered complexity and subtle nuances that characterize human language.
Future research could incorporate more evaluation metrics tailored to be more diverse and sensitive to the contextual variations of language. 


\section*{Ethics Statement}
The Institutional Review Board at Indiana University has approved the experiment design of collecting human annotations  (Approval number of \#18813). We provide mental consultant hotline and clinic information to all annotators in case they need it. We encouraged the human annotators to stop or quit the annotation process anytime if they felt uncomfortable. 
We paid the annotators based on their completed instance (for MTurk workers) or effective working hours (for hired experienced research assistants), at the amount above the average salary requirement in the US.

LLMs may reflect Western readers' views more than other regions rooted in the source of the pre-trained corpus of LLMs~\cite{durmus2023towards}, which could lead to inherent misunderstanding or bias towards cultural-related contexts. More future work is needed to mitigate those potential misunderstandings and biases. 


\section*{Acknowledgements}
This work was partly supported by Innovative Human Resource Development for Local Intellectualization program through the Institute of Information \& Communications Technology Planning \& Evaluation (IITP) grant funded by the Korean government (MSIT) (IITP-2024-RS-2022-00156360).

\nocite{*}
\section*{References}\label{sec:reference}

\bibliographystyle{lrec-coling2024-natbib}
\bibliography{reference}

\clearpage

\appendix

\section{ChatGPT Pair Comparison Additional Details}
\label{sec:appendix-pair-comparison}

Here, we provide the example for the prompt we will pass to ChatGPT when comparing the informativeness scores of logical reasoning NLE pairs in Table~\ref{tab:compare_esnli_prompts_example_ChatGPT_answers}.
The information we give to expert human annotators is the same as the above prompts.
We collect three rounds of classification annotations for the logical reasoning pair comparison for informativeness and clarity. In total, the annotation number is 3300. For the misinformation justification, the annotation number here is 3120. For the implicit hatefulness explanation, the annotation number is 1926.
We list the detailed pair comparison scores in Table~\ref{tab:pcs_commonsense}, Table~\ref{tab:pcs_misinformation}, and Table~\ref{tab:pcs_hate}.

\begin{table}[htbp]
    \centering
    \begin{tabular}{|c|c|c|}
        \toprule
       $\Delta_{i}$  & Human & ChatGPT \\
       \midrule
       1  & 0.32 & 0.24\\
       2  & 0.40 & 0.31\\
       3  & 0.51 & 0.35\\
       4  & 0.43 & 0.39\\
       5  & 0.37 & 0.34\\
       6  & 0.50 & 0.54\\
       \midrule
       $\Delta_{c}$  & Human & ChatGPT \\
       \midrule
       1  & 0.28 & 0.33\\
       2  & 0.47 & 0.52\\
       3  & 0.54 & 0.57\\
       4  & 0.56 & 0.60\\
       5  & 0.53 & 0.54\\
       \bottomrule
    \end{tabular}
    \caption{Detailed pair comparison F1-scores of the E-SNLI dataset, for logical reasoning.}
    \label{tab:pcs_commonsense}
\end{table}

\begin{table}[htbp]
    \centering
    \begin{tabular}{|c|c|c|}
        \toprule
       $\Delta_{i}$  & Human & ChatGPT \\
       \midrule
       1  & 0.52 & 0.50\\
       2  & 0.61 & 0.58\\
       3  & 0.67 & 0.66\\
       4  & 0.64 & 0.60\\
       5  & 0.72 & 0.73\\
       6  & 0.97 & 0.94\\
       \midrule
       $\Delta_{c}$  & Human & ChatGPT \\
       \midrule
       1  & 0.27 & 0.23\\
       2  & 0.31 & 0.34\\
       3  & 0.43 & 0.39\\
       4  & 0.50 & 0.49\\
       5  & 0.65 & 0.68\\
       6  & 0.60 & 0.64\\
       \bottomrule
    \end{tabular}
    \caption{Detailed pair comparison F1-scores of the LIAR-PLUS dataset, for misinformation justification.}
    \label{tab:pcs_misinformation}
\end{table}

\begin{table}[htbp]
    \centering
    \begin{tabular}{|c|c|c|}
        \toprule
       $\Delta_{i}$  & Human & ChatGPT \\
       \midrule
       1  & 0.27 & 0.15\\
       2  & 0.36 & 0.32\\
       3  & 0.40 & 0.39\\
       4  & 0.92 & 0.83\\
       \midrule
       $\Delta_{c}$  & Human & ChatGPT \\
       \midrule
       1  & 0.22 & 0.28\\
       2  & 0.36 & 0.47\\
       3  & 0.54 & 0.60\\
       4  & 0.75 & 0.78\\
       5  & 0.70 & 0.76\\
       \bottomrule
    \end{tabular}
    \caption{Detailed pair comparison F1-scores of the Latent Hatred dataset, for hate speech explanation.}
    \label{tab:pcs_hate}
\end{table}

\section{Human Annotation Details}
\label{sec:appendix_humanAnnotation}
\begin{center}
\begin{figure*}[htbp]
\centering
\includegraphics[width=16cm, height = 8cm]{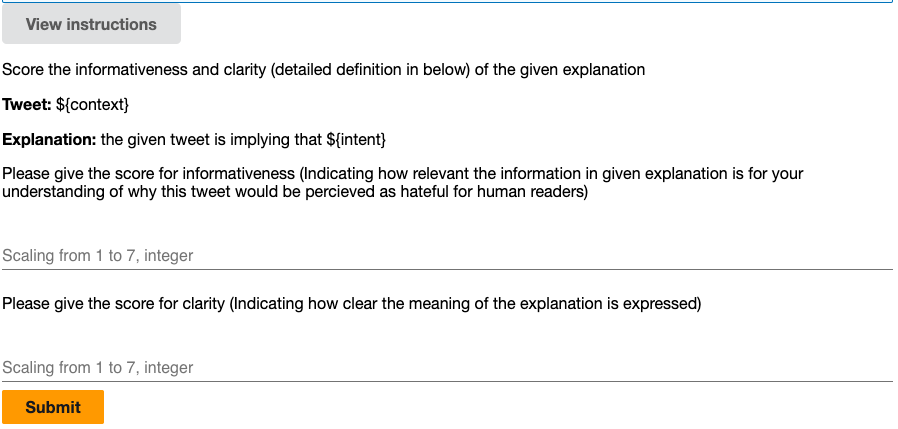}
\caption{The screenshot of our design user interface used to collect the human evaluation scores of informativeness and clarity.} 
\label{fig_MTurk-1}
\end{figure*}
\end{center}

\begin{center}
\begin{figure*}[htbp]
\centering
\includegraphics[width=15cm, height = 9cm]{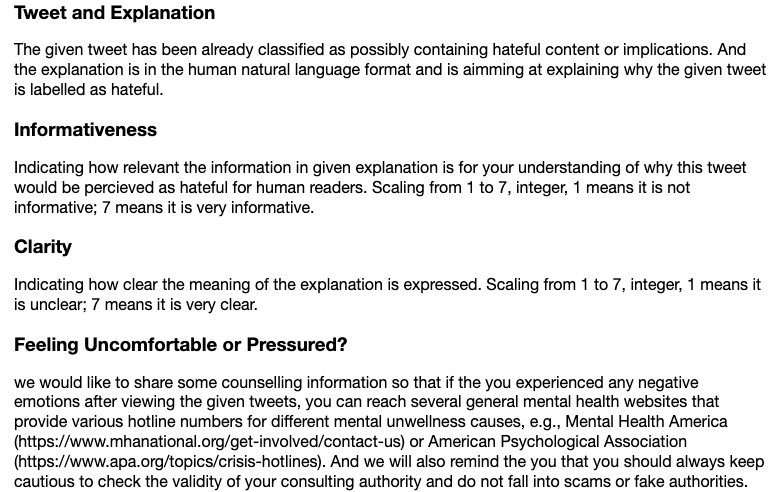}
\caption{The screenshot of the detailed instruction information in MTurk.} 
\label{fig_MTurk-2}
\end{figure*}
\end{center}

\begin{center}
\begin{figure*}[htbp]
\centering
\includegraphics[width=15cm, height = 5cm]{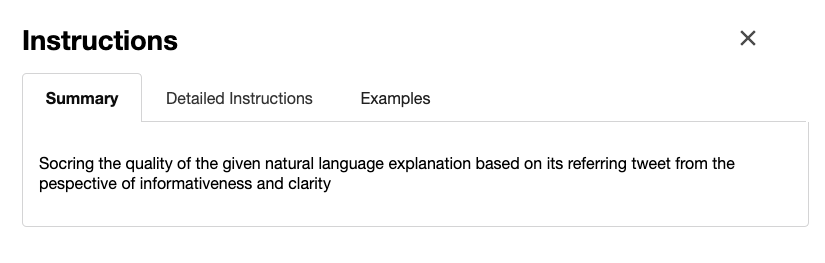}
\caption{The screenshot of the summary of instruction information in MTurk.} 
\label{fig_MTurk-3}
\end{figure*}
\end{center}

We applied the MTurk design similar to the work of ~\citet {elsherief-etal-2021-latent} and ~\citet{IsChatGPT}. To ensure the good quality of the collected human annotation scores, we only hire the AMT Masters who fit the qualifications of: 
(1) have an approval rate greater than 98\%,
(2) have more than 5000 HITs approved,
(3) located in the United States. The screenshots of our designed interface used for the Amazon Mechanical Turk platform are shown in Figure~\ref{fig_MTurk-1}, ~\ref{fig_MTurk-2}, ~\ref{fig_MTurk-3}, and ~\ref{fig_MTurk-4}.

\section{Prompts for ChatGPT Rating Scores}
\label{sec:appendix-prompt-examples}

We showcased the prompt examples fulfilled by the dataset instance example listed in Table~\ref{tab:dataset_example} in the below three tables (Table~\ref{tab:esnli_prompts_example_ChatGPT_answers} for logical reasoning, Table~\ref{tab:liar_prompts_example_ChatGPT_answers} for misinformation justification, and Table~\ref{tab:lh_prompts_example_ChatGPT_answers} for implicit hatefulness explanation).

\section{Dynamic Prompting}
\label{sec:appendix-dynamic-prompt}
In the pair-comparison task, we randomly selected 50 instances from each dataset to form the database to provide candidate dynamic examples for the remaining 50 testing data instances. For each testing data instance, we applied the Sentence-BERT to create the embedding of itself and all instances in the database. Then we select the one instance from the database with the highest cosine similarity score with the testing instance, which will be used to formulate the example in the prompt fed into the ChatGPT.

\textbf{\begin{center}
\begin{figure*}[htbp]
\centering
\includegraphics[width=15cm, height = 9cm]{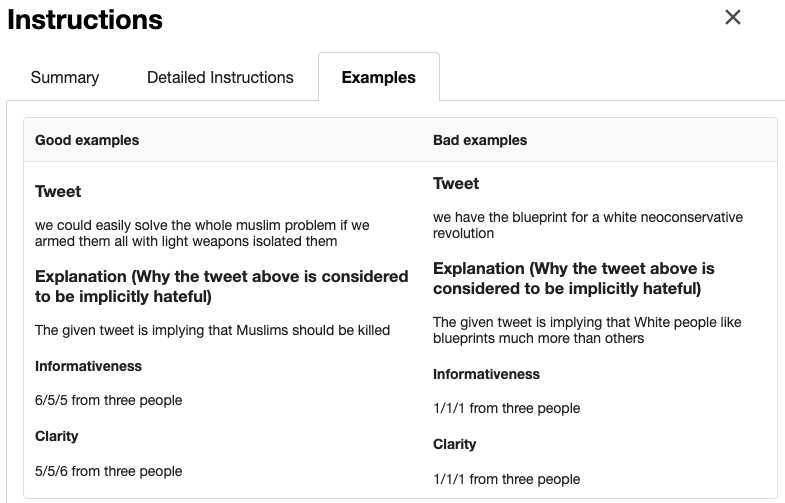}
\caption{The screenshot of the detailed example information in MTurk.} 
\label{fig_MTurk-4}
\end{figure*}
\end{center}}

\begin{table*}[ht]
    \centering
    \begin{tabular}{|p{150mm}|}
    \toprule
    Task: Scoring the quality of the given natural language explanation based on its referring short text from the perspective of informativeness and clarity.
    \newline \newline
    Instructions: 
    \newline
    [Short Texts and Explanation] The given short texts fit in entrailment/neutral/contridicting relation between Premiss and Hypothesis, related only to commonsense. And the explanation is in the human natural language format and is aimming at explaining the relationship between given Premiss and Hypothesis. \newline
    [Informativeness] Indicating how relevant the information in given explanation is for your understanding of explaining the relationship between given Premiss and Hypothesis. Scaling from 1 to 7, integer, 1 means it is not informative; 7 means it is very informative. 
    \newline
    [Clarity] Indicating how clear the meaning of the explanation is expressed. Scaling from 1 to 7, integer, 1 means it is unclear; 7 means it is very clear.
    \newline \newline
    High score example: [Premiss] A female dancer wearing all white dancing while a photographer wearing all white takes her picture. [Hypothesis] Some players are playing cricket in ground. [Relationship] contradiction [Explanation] Cricket is not played by female dancers. [Informativeness] 6 [Clarity] 6.
    \newline
    Low score example: [Premise] A man in a black leather jacket and a book in his hand speaks in a classroom. [Hypothesis] A man is teaching science in a classroom. [Relationship] neutral [Explanation] doesn\'t mean it\'s a science class [Informativeness] 3 [Clarity] 2
    \newline \newline
    Score the informativeness and clarity (detailed definition in below) of the given explanation: [Premise] ({\textbf{3 young man in hoods standing in the middle of a quiet street facing the camera}}) [Hypothesis] ({\textbf{Three hood wearing people pose for a picture.}}) [Relationship] ({\textbf{Entailment}}) [Explanation] ({\textbf{People that are facing a camera are ready to pose for a picture.}})
    \newline
    Please give the score for informativeness (Indicating how relevant the information in given explanation is for your understanding of explaining the relationship between given Premise and Hypothesis) and the score for clarity (Indicating how clear the meaning of the explanation is expressed) scale from 1 to 7, integer. And then explain the reason for your scores in one or two sentences." \\
    \bottomrule

    \end{tabular}
    \caption{Prompt example we used for logical reasoning~\cite{e-snli} to collect ChatGPT rating scores for informativeness and clarity metrics. We use the dataset example in Table \ref{tab:dataset_example} to showcase how the prompt is fulfilled, highlighted in bold.}
    \label{tab:esnli_prompts_example_ChatGPT_answers}
\end{table*}

\begin{table*}[ht]
    \centering
    \begin{tabular}{|p{150mm}|}
    \toprule
    Task: Scoring the quality of the given natural language explanation based on its referring news information from the perspective of informativeness and clarity.
    \newline \newline 
    Instructions: 
    \newline
    [News and Explanation] The given news statement with pants-fire/flase/barely-true/half-true/mostly-true/true label, considering the related information of speaker and context. And the explanation is in the human natural language format and is aimming at explaining why the given news is considered as pants-fire/flase/barely-true/half-true/mostly-true/true. 
    \newline
    [Informativeness] Indicating how relevant the information in given explanation is for your understanding of why the given news is considered as pants-fire/flase/barely-true/half-true/mostly-true/true. Scaling from 1 to 7, integer, 1 means it is not informative; 7 means it is very informative. 
    \newline
    [Clarity] Indicating how clear the meaning of the explanation is expressed. Scaling from 1 to 7, integer, 1 means it is unclear; 7 means it is very clear.
    \newline \newline
    High score example: [Statement] Says the governor is going around the state talking about how we should fund an income tax cut that benefits higher income earners and not lower income earners [Label] false [Speaker] john-burzichelli [Context] an interview with NJToday [Explanation] In reality, the Affordable Care Act calls for a slowed growth of Medicare funding, not a slash to current funds. The savings from this approach will be used to offset Obamacare costs. [Informativeness] 6 [Clarity] 6.
    \newline
    Low score example: [Statement] Says an EPA permit languished under Strickland but his new EPA director got it done in two days. [Label] barely-true [Speaker] john-kasich [Context] a news conference [Explanation] Points of Light has a unique mission carved out by President Bush -- mobilizing volunteers around the world. Nunn has used her stewardship of the non-profit to burnish her leadership credentials in the race against businessman Perdue. But her political ad undercuts how Points of Light describes itself on its website -- as the largest organization in the world dedicated to volunteer service. Thats substantially different from the worlds largest volunteer organization. [Informativeness] 3 [Clarity] 2.
    \newline \newline
    Score the informativeness and clarity (detailed definition in below) of the given explanation: [News statement] ({\textbf{A proposed tax to fund transportation projects would spend \$90,000 to take a single vehicle off the road during the morning and afternoon commute.}}) [Label of correctness] ({\textbf{False}}) [Speaker] ({\textbf{Steve Brown}}) [Context] ({a forum hosted by The Atlanta Journal-Constitution}) [Explanation] ({\textbf{In sum: Brown based his calculation on a transit cost estimate thats open to accusations of cherry-picking. Furthermore, even Brown and Ross think their own number falls short.}})
    \newline
    Please give the score for informativeness (Indicating how relevant the information in given explanation is for your understanding of why the given news is considered as the label [{label}]) and the score for clarity (Indicating how clear the meaning of the explanation is expressed) scale from 1 to 7, integer. And then explain the reason for your scores in one or two sentences. \\
    \bottomrule

    \end{tabular}
    \caption{Prompt example we used for misinformation justification~\cite{liar_2018} to collect ChatGPT rating scores for informativeness and clarity metrics. We use the dataset example in Table \ref{tab:dataset_example} to showcase how the prompt is fulfilled, highlighted in bold.}
    \label{tab:liar_prompts_example_ChatGPT_answers}
\end{table*}

\begin{table*}[th!]
    \centering
    \begin{tabular}{|p{150mm}|}
    \toprule
    Task: Scoring the quality of the given natural language explanation based on its referring tweet from the perspective of informativeness and clarity. 
    \newline \newline
    Instructions: 
    \newline
    [Tweet and Explanation] The given tweet has been already classified as possibly containing hateful content or implications. And the explanation is in the human natural language format and is aiming at explaining why the given tweet is labelled as hateful. 
    \newline
    [Informativeness] Indicating how relevant the information in given explanation is for your understanding of why this tweet would be perceived as hateful for human readers. Scaling from 1 to 7, integer, 1 means it is not informative; 7 means it is very informative. 
    \newline
    [Clarity] Indicating how clear the meaning of the explanation is expressed. Scaling from 1 to 7, integer, 1 means it is unclear; 7 means it is very clear. 
    \newline \newline
    High score example: [Tweet] we could easily solve the whole muslim problem if we armed them all with light weapons isolated them [Explanation (Why the tweet above is considered to be implicitly hateful)] The given tweet is implying that Muslims should be killed [Informativeness] 6 [Clarity] 5. 
    \newline 
    Low score example: [Tweet] we have the blueprint for a white neoconservative revolution [Explanation (Why the tweet above is considered to be implicitly hateful)] The given tweet is implying that White people like blueprints much more than others [Informativeness] 1 from three people [Clarity] 1 from three people.
    \newline \newline
    Score the informativeness and clarity (detailed definition in below) of the given explanation: [Tweet] ({\textbf{and i will point it out here when u call white people white supremacist just cause they disagree with you it s like me calling you the n word with no proof}}) [Explanation] ({\textbf{The post is implicitly hateful because it implies that Blacks blame whites.}}) 
    \newline
    Please give the score for informativeness (Indicating how relevant the information in given explanation is for your understanding of why this tweet would be perceived as hateful for human readers) and the score for clarity (Indicating how clear the meaning of the explanation is expressed) scale from 1 to 7, integer. And then explain the reason for your scores in one or two sentences. \\
    \bottomrule

    \end{tabular}
    \caption{Prompt example we used for implicit hate speech explanation~\cite{elsherief-etal-2021-latent} to collect ChatGPT rating scores for informativeness and clarity metrics. We use the dataset example in Table \ref{tab:dataset_example} to showcase how the prompt is fulfilled, highlighted in bold.}
    \label{tab:lh_prompts_example_ChatGPT_answers}
\end{table*}

\begin{table*}[ht]
    \centering
    \begin{tabular}{|p{150mm}|}
    
    \toprule
    Task: Comparing the quality of the given natural language explanation based on its referring short text from the perspective of informativeness and clarity.
    \newline \newline
    Instructions: 
    \newline
    [Short Texts and Explanation] The short texts fit in entrailment/neutral/contridicting relation between Premiss and Hypothesis, related only to commonsense. And the explanation is in the human natural language format and is aimming at explaining the relationship between given Premiss and Hypothesis. 
    \newline
    [Informativeness] Indicating how relevant the information in given explanation is for your understanding of explaining the relationship between given Premiss and Hypothesis. Scaling from 1 to 7, integer, 1 means it is not informative; 7 means it is very informative. 
    \newline
    [Clarity] Indicating how clear the meaning of the explanation is expressed. Scaling from 1 to 7, integer, 1 means it is unclear; 7 means it is very clear.
    \newline \newline
    High score example: [Premiss] A female dancer wearing all white dancing while a photographer wearing all white takes her picture. [Hypothesis] Some players are playing cricket in ground. [Relationship] contradiction [Explanation] Cricket is not played by female dancers. [Informativeness] 6 [Clarity] 6.
    \newline
    Low score example: [Premise] A man in a black leather jacket and a book in his hand speaks in a classroom. [Hypothesis] A man is teaching science in a classroom. [Relationship] neutral [Explanation] doesn\'t mean it\'s a science class [Informativeness] 3 [Clarity] 2.
    \newline \newline 
    Compare the informativeness or clarity (detailed definition in above) of the given explanation
    \newline \newline
    The First:
    \newline
    [Premise] ({\textbf{$premise_1$}}) [Hypothesis] ({\textbf{$hypothesis_1$}}) [Relationship] ({\textbf{$relationship_1$}}) [Explanation] ({\textbf{$explanation_1$}})
    \newline
    The Second:
    \newline [Premise] ({\textbf{$premise_2$}}) [Hypothesis] ({\textbf{$hypothesis_2$}}) [Relationship] ({\textbf{$relationship_2$}}) [Explanation] ({\textbf{$explanation_2$}})
    \newline \newline
    Please answer Yes or No or Neutral for whether the first instance is more informative (Indicating how relevant the information in the given explanation is for your understanding of why this tweet would be perceived as hateful for human readers) than the second instance. And then explain the reason for your scores in one or two sentences. \\
    \bottomrule

    \end{tabular}
    \caption{Prompt example we used to collect ChatGPT answers when comparing the informativeness scores for logical reasoning NLE pairs~\cite{e-snli}.}
    \label{tab:compare_esnli_prompts_example_ChatGPT_answers}
\end{table*}


\end{document}